\documentclass[autodetect-engine,dvipdfmx,10pt,a4paper,twocolumn]{article}

\usepackage[linesnumbered,ruled]{algorithm2e}
\usepackage{bm}
\usepackage{amsmath}
\usepackage{amssymb}
\usepackage{amsthm}
\usepackage{color}
\usepackage{dsfont}
\usepackage{booktabs}
\usepackage{framed}
\usepackage{tabularx}
\usepackage{graphicx}

\newenvironment{tsaligned}{%
\begin{equation}\begin{aligned}%
}{%
\end{aligned}\end{equation}%
}
\newenvironment{tsaligned*}{%
\begin{equation*}\begin{aligned}%
}{%
\end{aligned}\end{equation*}%
}

\theoremstyle{definition}
\newtheorem{theorem}{Theorem}
\theoremstyle{definition}
\newtheorem{lemma}[theorem]{Lemma}
\theoremstyle{definition}

\theoremstyle{definition}

\theoremstyle{definition}

\theoremstyle{definition}
\newtheorem{definition}[theorem]{Definition}
\theoremstyle{definition}
\newtheorem{proposition}[theorem]{Proposition}
\theoremstyle{definition}
\newtheorem{property}[theorem]{Property}
\theoremstyle{definition}
\newtheorem{example}[theorem]{Example}
\theoremstyle{definition}
\newtheorem{conjecture}[theorem]{Conjecture}
\theoremstyle{definition}

\theoremstyle{definition}

\theoremstyle{definition}
\newtheorem{corollary}[theorem]{Corollary}
\theoremstyle{definition}
\newtheorem{exercise}[theorem]{Exercise}
\theoremstyle{definition}
\newtheorem{simulation}[theorem]{Simulation}

{\endMakeFramed}

\newenvironment{greenleftbar}{%
\MakeFramed {\advance\hsize-\width \FrameRestore}}%
{\endMakeFramed}

\newenvironment{lightgrayleftbar}{%
\MakeFramed {\advance\hsize-\width \FrameRestore}}%
{\endMakeFramed}

\newenvironment{example-waku}
{\begin{lightgrayleftbar}\begin{example}}
{\end{example}\end{lightgrayleftbar}}
\newenvironment{exercise-waku}
{\begin{lightgrayleftbar}\begin{exercise}}
{\end{exercise}\end{lightgrayleftbar}}
\newenvironment{simulation-waku}
{\begin{greenleftbar}\begin{simulation}}
{\end{simulation}\end{greenleftbar}}
\newenvironment{proposition-waku}
{\begin{oframed}\begin{proposition}}
{\end{proposition}\end{oframed}}
\newenvironment{definition-waku}
{\begin{oframed}\begin{definition}}
{\end{definition}\end{oframed}}
\newenvironment{lemma-waku}
{\begin{oframed}\begin{lemma}}
{\end{lemma}\end{oframed}}
\newenvironment{theorem-waku}
{\begin{oframed}\begin{theorem}}
{\end{theorem}\end{oframed}}
\newenvironment{property-waku}
{\begin{oframed}\begin{property}}
{\end{property}\end{oframed}}
\newenvironment{corollary-waku}
{\begin{oframed}\begin{corollary}}
{\end{corollary}\end{oframed}}
\newenvironment{conjecture-waku}
{\begin{oframed}\begin{conjecture}}
{\end{conjecture}\end{oframed}}

\newcommand{\0}{{\bm{0}}}

\newcommand{\va}{{\bm{a}}}
\newcommand{\vb}{{\bm{b}}}

\newcommand{\vt}{{\bm{t}}}

\newcommand{\vv}{{\bm{v}}}
\newcommand{\vw}{{\bm{w}}}
\def\x{\bm{x}}

\newcommand{\y}{{\bm{y}}}

\newcommand{\vA}{{\bm{A}}}

\newcommand{\vB}{{\bm{B}}}

\newcommand{\bE}{{\mathbb{E}}}

\newcommand{\cE}{{\mathcal{E}}}

\newcommand{\cF}{{\mathcal{F}}}
\newcommand{\vG}{{\bm{G}}}

\newcommand{\cH}{{\mathcal{H}}}

\newcommand{\vI}{{\bm{I}}}

\newcommand{\cL}{{\mathcal{L}}}

\newcommand{\cN}{{\mathcal{N}}}

\newcommand{\cQ}{{\mathcal{Q}}}

\newcommand{\cP}{{\mathcal{P}}}

\newcommand{\bR}{{\mathbb{R}}}

\newcommand{\vW}{{\bm{W}}}

\newcommand{\Y}{{\bm{Y}}}

\newcommand{\vthet}{{\bm{\theta}}}

\newcommand{\argmax}{\mathop{\textrm{argmax}}\limits}

\newcommand{\tsrev}[1]{\textcolor{black}{#1}}
\newcommand{\tsrevtwo}[1]{\textcolor{black}{#1}}

\newcommand{\tslong}[1]{{#1}}
\newcommand{\tsshort}[1]{{}}

\begin{document}
\title{Multi-Target Tobit Models for Completing Water Quality Data}

\author{Yuya Takada${}^{1}$ and Tsuyoshi Kato${}^{1}$
\\
\normalsize
${}^1$~Faculty of Science and Technology, Gunma University, \\
\normalsize
Tenjin-cho 1-5-1, Kiryu, Gunma 376-8515, Japan. %
}
\maketitle
\begin{abstract}
  Monitoring microbiological behaviors in water is crucial to manage public health risk from waterborne pathogens, although quantifying the concentrations of microbiological organisms in water is still challenging because concentrations of many pathogens in water samples may often be below the quantification limit, producing censoring data. To enable statistical analysis based on quantitative values, the true values of non-detected measurements are required to be estimated with high precision. Tobit model is a well-known linear regression model for analyzing censored data. One drawback of the Tobit model is that only the target variable is allowed to be censored. In this study, we devised a novel extension of the classical Tobit model, called the \emph{multi-target Tobit model}, to handle multiple censored variables simultaneously by introducing multiple target variables. For fitting the new model, a numerical stable optimization algorithm was developed based on elaborate theories. Experiments conducted using several real-world water quality datasets provided an evidence that estimating multiple columns jointly gains a great advantage over estimating them separately. 
\end{abstract}

  Keywords: 
  water quality data, 
  censoring, 
  data imputation, 
  Tobit model, 
  probabilistic model, 
  linear regression. 
\section{Introduction}
Water is one of the most critical resources for life, although pollution of water has threatened public health safety. Even clear water that looks drinkable may be polluted with microbiological organisms leading to serious health risk. Exposure to waterborne pathogens may cause pernicious diarrheal diseases estimated to kill 485,000 people each year~\cite{WHO2022a}. Waterborne outbreaks may damage not only the individual finances, but also the national economy~\cite{Bridle20a}. For these reasons, managing public health risks caused by microbiological water contamination has remained a major concern around the globe. 

Regulatory agencies have responsibility for development of water quality guidelines to protect public health~\cite{WadPaiEis03,WHO2022a}.
The guidelines are designed to ensure health safety with minimal monitoring costs. Direct measurement of waterborne pathogenic factors is too costly to be practical to incorporate into routine monitoring. The behavior of aquatic pathogenic microorganisms is influenced by multiple influences such as hydrogeology, temperature, and season. Therefore, it is necessary to observe the behavior of microorganisms in the water in each environment and determine environmental standards and monitoring processes appropriately according to the cost and economic capacity of the community.

Quantifying microbial concentrations in water is challenging because concentrations of many pathogens in water samples may often be below the quantification limit, producing censoring data. To enable statistical analysis based on quantitative values, such as correlation analysis, the true values of non-detected measurements are required to be estimated with high precision. 

Tobit model~\cite{Amemiya1984,KatKobOis19} is known as an extension of a linear regression method for analyzing censored data consisting of a target variable and explanatory variables. 
Only the target variables are allowed to be censored, and the numerical values of all the explanatory variables must be given for fitting of Tobit model (Figure~\ref{fig:grphmdl}(a),(b)). 
However, for analysis of relationship between pathogenic concentrations, multiple censored variables are often contained in a microbiological water quality dataset. 
An approach to complete microbiological water quality data is changing the target variable to each censored pathogenic concentration variable and repeatedly applying the Tobit model. 
For this approach to work, however, the non-detected pathogen concentrations used for the explanatory variables must be filled in with some value before it is applied. 
This approach is eclectic and sub-optimal to estimation of multiple censored pathogenic concentrations. 

In this study, we devised a novel extension of the classical Tobit model to handle multiple censored variables simultaneously by introducing multiple target variables (Figure~\ref{fig:grphmdl}(c),(d)). The extended model is referred to as the \emph{multi-target Tobit model}~(MTTM). The reformulation is based on the fact that a term in the likelihood function of Tobit model can be expressed with the minimum of a Kullback-Leibler 
\tsrev{%
divergence
}%
(Theorem~\ref{thm:sttm-is-max-F}). The proposed extension of Tobit model was achieved by combining multiple likelihood functions, with a theoretical justification offered by Theorem~\ref{thm:sttm-is-max-F}. The main discovery of this study is as follows: 
\begin{itemize}
\item
Each step can be expressed in a closed form when the block-coordinate ascent method is used for maximizing the superimposed objective function (Theorem~\ref{thm:ca-m-upd}). 
\end{itemize}
This finding helped development of an efficient optimization algorithm for maximizing the reformulated objective function. With help of this, neither hyper-parameter nor integral computation over multidimensional spaces is required for data fitting, which enables numerically stable optimization. Computational experiments using real-world microbiological water quality data empirically showed that the novel reformulated model could has significantly higher accuracy than the classical Tobit model in estimating the censored pathogenic concentrations.
\tsrev{%
Although this paper is written primarily for application to water quality engineering, the proposed method is more general in scope and can be used to supplement a wide variety of censored data. Possible extensions and applicability are discussed in Section~\ref{s:discuss}.
}%
\section{Related work}
\label{s:related}

To confront frequent waterborne illness outbreaks occurring worldwide, tremendous efforts have been devoted for elucidation of the relationships between the concentrations of various viruses and bacteria in water~\cite{SaxBhaKai15}. Analyses of these relationships are used to identify factors that control the persistence and distribution of pathogenic microorganisms in specific environments and ecosystems and to understand the potentials and the limitations of indicator microorganisms. For example, Anderson-Coughlin et al explored the relationships among viruse contaminations measured in a 17-month monitoring study of human enteric viruses and indicators in surface and reclaimed water in the central Atlantic region of the United States~\cite{AndCraKel21}. They used real-time quantitative PCR to determine the concentrations of aichi virus, hepatitis A virus, norovirus GI, and norovirus GII, and reported that PMMoV and enteric viruses are significantly correlated at the detection level, and salinity is significantly correlated to the detection of enteric viruses and PMMoV. Farkas et al performed meta-analysis based on the data reported in 127 peer-reviewed papers, and found that the human mastadenovirus genus has great potential for measuring sewage contamination~\cite{FarWalAdr20}. Rochelle-Newall et al discussed the factors controlling fecal indicator bacteria in temperate zones and their applicability to tropical environments~\cite{RocNguLe15}. Developing quantification techniques for many waterborne pathogens is an ongoing process, still producing left-censoring for pathogenic concentrations in water due to high detection limits~\cite{SaxBhaKai15}. Many studies including these studies \cite{SaxBhaKai15,AndCraKel21,FarWalAdr20,RocNguLe15} have still either discarded the quantitative values and analyzed the data as binary data, or analyzed only the quantifiable data. Binarized data would result in a significant loss of information, and analysis based solely on quantifiable data would lead to bias in estimation. Tobit approach offers a theoretically valid framework for analysis of data with censoring. 

Tobit models have been used long as a statistical tool in economics and social sciences, including labor economics, consumer behavior, financial asset selection, and demand functions for consumer durable goods for the purpose of analyzing data including corner solutions~\cite{McDonald80}. 
In recent years, the potential of the Tobit model has been recognized in many academic fields and has spread to a vast array of applications including nursing study~\cite{LinChe11}, healthcare study~\cite{SayWhiJud20}, analysis on survival times~\cite{QiGong18}, traffic accidents~\cite{RongjieYu20}, tax information~\cite{ZijingWang22}, school ranking~\cite{Kirjavainen97}, global financial crisis~\cite{Amirthalingam2016}, and employment efficacy of graduates~\cite{FengnaZhang21}. Such recent studies introduced more complex variants of Tobit models including bayesian selection Robit model~\cite{PengDing14}, multilevel Tobit model~\cite{SayWhiJud20}, bayesian random-parameter model~\cite{YanyongGuo2019}, SUR model~\cite{RiverHuang99}, random effect Tobit model~\cite{WeiWang16}, mixed effect Tobit model~\cite{Dagne12}. If the model has only two variates, data fitting can be performed typically with a simple computation~\cite{Blundell87,SongnianChen11}, but many models with three or more variates necessarily resort to computationally heavy multiple integral calculations~\cite{Pakman13}. The algorithm introduced in this study for water quality data completion does not require integration over multidimensional spaces, and each iteration consists of only simple calculations, thus ensuring numerical stability. 

\section{Preliminaries}
This paper deals with probability density functions such that the densities exactly vanish on some ray~$[u,+\infty)$. For such density functions $p,q:\bR\to\bR$, the entropy and the Kullback-Leibler divergence, respectively, are defined as
\begin{tsaligned}
    & \cH[q]
    :=
    -\int_{-\infty}^{u} q(x) \log q(x) dx 
\end{tsaligned}  
and
\begin{tsaligned}  
    & \text{KL}[q||p] 
    :=
    -\cH[q] - \int_{-\infty}^{u} q(x) \log p(x) dx.  
\end{tsaligned}

\begin{figure}[t!]
  \begin{center}
    \begin{tabular}{ll}
      (a) & (b) STTM
      \\
      \includegraphics[width=0.13\textwidth]{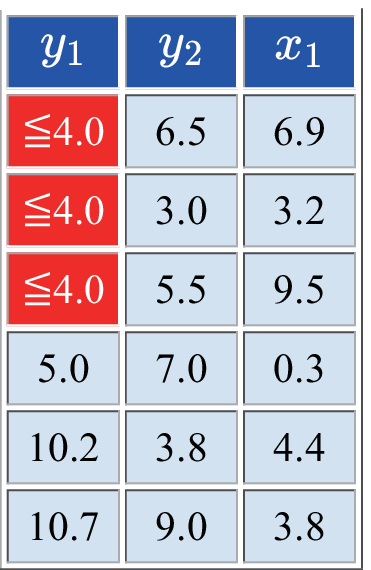}
      &
      \includegraphics[width=0.18\textwidth]{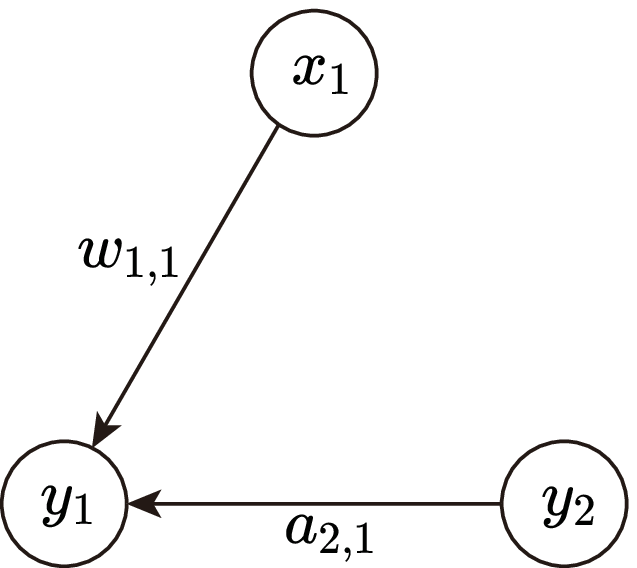}
      \\
      \\
      (c) & (d) MTTM
      \\
      \includegraphics[width=0.13\textwidth]{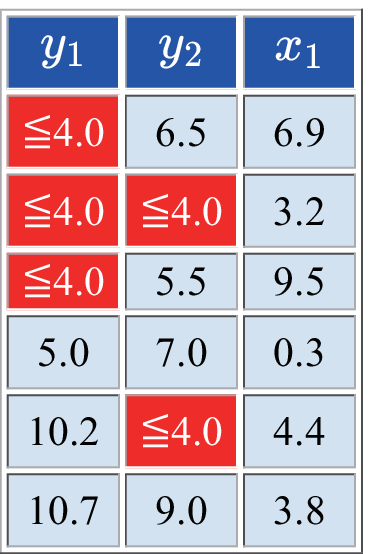}
      &
      \includegraphics[width=0.18\textwidth]{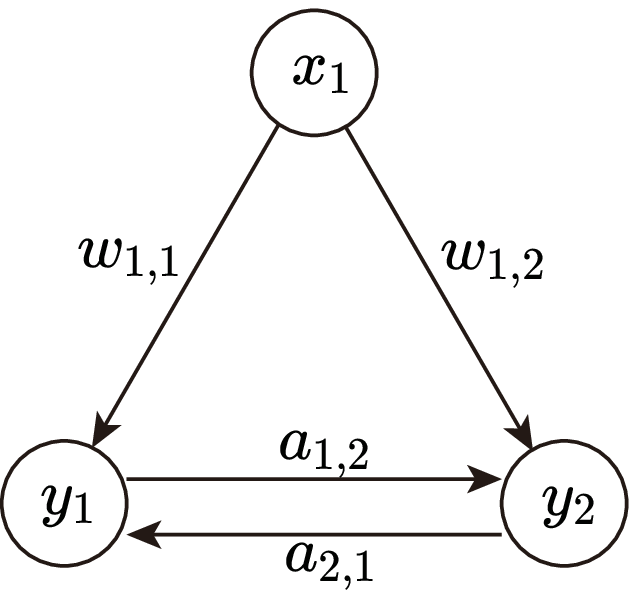}
    \end{tabular}
  \end{center}
  \caption{%
    \textbf{Classical and new Tobit models. }
    The classical Tobit model, STTM, is a framework that fits a regression model (see (b)) from a table containing 
    \tsrev{%
    a column 
    }
    with censoring (see (a)), where the target variable is the column with censoring. 
    The new model, MTTM, can jointly fit more than one linear regression model (see (d)) from a table containing multiple censoring columns (see (c)) by 
    \tsrev{%
    combining
    }
    these regression models. 
    \tsrev{%
    Therein, $a_{1,2}$, $a_{2,1}$, $w_{1,2}$ and $w_{2,1}$ are regression coefficients. 
    }%
    \label{fig:grphmdl}
  }
\end{figure}

\section{Single target Tobit model (STTM) }
\label{s:sttm}
The classical Tobit model~\cite{Amemiya1984} assumes the linear relationship
\begin{tsaligned*}
   y_{i} = \left<\vw,\x_{i}\right> + \epsilon_{i}
\end{tsaligned*}
given a dataset~$(\x_{i},y_{i})\in\bR^{d}\times\bR$ 
$(i\in [n]:=\{1,\dots,n\})$, 
and determines the partial regression coefficient vector $\vw\in\bR^{d}$, 
where $\epsilon_{i}$ is a normal noise, $\epsilon_{i}\sim\cN(0,\beta^{-1})$. 
Let $u_{}$ be the \emph{detection limit}, and suppose that $n_{\text{v}}$ target data are over the detection limit. 
\tsrev{%
To speak more precisely, the variable $u$ should be called the \emph{quantification limit}~\cite{KatKobOis19}. 
}%
Without loss of generality, the indices of examples can be exchanged so that 
$y_{i} > u_{}$ for $i\in[n_{\text{v}}]:=\{1,\dots,n_{\text{v}}\}$ and 
$y_{i} \le u_{}$ for $i\in[n]\setminus [n_{\text{v}}] = \{n_{\text{v}}+1,\dots,n\}$. 
For data below the detection limit, in contrast to the vanilla linear regression analysis, the only information available for estimating the regression coefficient vector is that the data are censored; the actual values of the target data cannot be accessed. 
For fitting the Tobit model to a censored dataset, the following log-likelihood function is used: 
\begin{tsaligned}\label{eq:L-s-def}
   &
   \cL_{\text{s}}(\vthet_{\text{s}})
   :=
   \sum_{i=1}^{n_{\text{v}}}\log\cN( y_{i} \,;\, \left<\vw,\x_{i}\right>, \beta^{-1} )
   \\
   &
   \quad
   +
   \sum_{i=n_{\text{v}}+1}^{n}\log
   \int_{-\infty}^{u_{}}\cN( t_{i} \,;\, \left<\vw,\x_{i}\right>, \beta^{-1} ) d t_{i},  
\end{tsaligned}
where $\vthet_{\text{s}}:=(\vw,\beta)$ is the model parameters. 
Newton method is often employed to find the optimal value of $\vthet_{\text{s}}$~\cite{Amemiya1984}. 

The negative of the second term in \eqref{eq:L-s-def} is described as the minimal Kullback-Leibler divergence between two 
\tsrev{%
probability
}%
density functions detailed later. 
In the next section, we shall demonstrate how this fact can be used to reformulate the classical Tobit model to handle multiple target variables.
Denote by $\cP$ the set of 
\tsrev{%
probability
}%
density functions $\bR\to\bR$, and let us introduce the following $n$ 
\tsrev{%
probability
}%
density functions $q_{1},\dots,q_{n}\in\cP$: The first $n_{\text{v}}$ density functions are defined using Dirac delta functions $\delta(\cdot)$: $\forall i\in[n_{\text{v}}],$
\begin{tsaligned*}
   q_{i}(t) = \delta(t-y_{i}). 
\end{tsaligned*}
Let $\cQ_{i}:=\{ \delta(\cdot-y_{i})\}$ for $i\in[n_{\text{v}}]$. 
The remaining $n_{\text{h}}(:=n-n_{\text{v}})$ density functions $q_{n_{\text{v}}+1},\dots,q_{n}\in\cP$ are assumed to satisfy that $\forall i\in[n]\setminus[n_{\text{v}}]$, 
\begin{tsaligned}\label{eq:qi-cond-stbt}
   \forall t > u, \quad q_{i}(t)=0. 
\end{tsaligned}
For $i\in[n]\setminus[n_{\text{v}}]$, let 
$\cQ_{i}:=\{ q_{i}\in\cP \,|\, \forall t> u, \, q_{i}(t) = 0 \}$
and 
$\cQ_{\text{s}}:=\cQ_{1}\times\dots\times\cQ_{n}$. 
For the second term in \eqref{eq:L-s-def}, it can be shown that
\begin{tsaligned}\label{eq:min-KL-over-Q-is-neg-Z-in-main}
&-\log
\int_{-\infty}^{u_{}}\cN( t_{i} \,;\, \left<\vw,\x_{i}\right>, \beta^{-1} ) d t_{i}
\\
& 
\qquad
=
\min_{q_{i}\in\cQ_{i}}
\text{KL}
\left[q_{i}||
\cN(\cdot\,;\,\left<\vw,\x_{i}\right>,\beta^{-1})
\right]. 
\end{tsaligned}
%
Derivation of the equation~\eqref{eq:min-KL-over-Q-is-neg-Z-in-main} is referred to \tsrevtwo{Appendix~\ref{s:proof-min-KL-over-Q-is-neg-Z-in-main}}. %
This fact immediately leads to the following theorem. 
\begin{theorem-waku}\label{thm:sttm-is-max-F}
   For any 
   $\vthet_{\text{s}}=(\vw,\beta)\in\bR^{d}\times\bR$
   such that $\beta>0$, 
   it holds that 
\begin{tsaligned*}
   \cL_{\text{s}}(\vthet_{\text{s}})
   =
   \max_{q\in\cQ_{\text{s}}}\cF_{\text{s}}(\vthet_{\text{s}},q)
\end{tsaligned*}
where $\cF_{\text{s}}(\vthet_{\text{s}},q)$ is defined as 
\begin{tsaligned*}
   \cF_{\text{s}}(\vthet_{\text{s}},q) &:= 
   \sum_{i=n_{\text{v}}+1}^{n}
   \cH[q_{i}]
   +
   \sum_{i=1}^{n}
   \int
   q_{i}(t_{i})\cdot
   \\
   &
   \qquad
   \log
   \cN( t_{i} \,;\, \left<\vw,\x_{i}\right>, \beta^{-1} ) d t_{i}
 \end{tsaligned*}
 for $q:=(q_{1},\dots,q_{n})\in\cQ_{\text{s}}$, 
and $\cH[\cdot]:\cP\to\bR$ denotes the entropy. 
\end{theorem-waku}
\tsrev{%
The second term of $\cF_{\text{s}}(\vthet_{\text{s}},q)$ corresponds to the expected complete-data log-likelihood function also known as the \emph{Q-function}~\cite{bishop06}. 
}%
The possibility of an approach for maximum likelihood estimation alternative to Newton method is implied by Theorem~\ref{thm:sttm-is-max-F}. 
The alternative maximum likelihood estimation approach finds the pair of the parameters~$\vthet_{\text{s}}$ and the function set~$q$ that jointly maximizes the functional value~$\cF_{\text{s}}(\vthet_{\text{s}},q)$ rather than directly maximizing the likelihood function $\cL_{\text{s}}(\vthet_{\text{s}})$ with respect to the model parameters~$\vthet_{\text{s}}$.

Using the model for analysis of microbiological water quality data encounters a crucial drawback: only a single variable can be censored. What brings this drawback is the nature of the model: only the target variable can have uncertainty. Hereinafter, the classical Tobit model shall be referred to as the \emph{single target Tobit model} (\textbf{STTM}). To address this shortcoming, a new formulation containing multiple target variables was devised. Theorem~\ref{thm:sttm-is-max-F} shall be used to justify the reformulation, as described in the next section. 

\section{Multi-target Tobit model (MTTM)}
\label{s:mttm}
In this section, a new extension of the classical model, MTTM is presented to directly involve multiple censored variables for fitting and imputation. 
Let $m$ be the number of target variables. The sample dataset is expressed as
\begin{tsaligned*}
   (\x_{1},\y_{1}),\dots,(\x_{n},\y_{n})\in\bR^{d}\times\bR^{m}. 
\end{tsaligned*}
Each example in the dataset consists of $d$-dimensional explanatory variable vector $\x_{i}\in\bR^{d}$ and $m$-dimensional target vector 
\begin{tsaligned*}
  \y_{i}:=\left[ y_{1,i}, \dots, y_{m,i} \right]^\top
\end{tsaligned*}
Denote by 
$\y_{\setminus k,i} := \left[ y_{1,i}, \dots, y_{k-1,i}, y_{k+1,i}, \dots, y_{m,i} \right]^\top$ 
the $(m-1)$-dimensional vector obtained by excluding the $k$-th entry from the target variable vector $\y_{i}\in\bR^{m}$. 
MTTM is a fusion of $m$ linear regression models: 
\begin{tsaligned*}
k\in[m],\quad
y_{k,i}= 
\left<\va_{k},\y_{\setminus k,i}\right> + 
\left<\vw_{k},\x_{i}\right> + \epsilon_{k,i}, 
\end{tsaligned*}
where $\va_{k}\in\bR^{m-1}$ and $\vw_{k}\in\bR^{d}$ are the partial regression coefficients of the $k$-th linear regression model; $\epsilon_{k,i}\sim\cN(0,\beta^{-1})$ is a zero-mean random noise with variance $\beta^{-1}$.  
Note that each target variable plays a role of an explanatory variable in other linear regression models. 

MTTM is fitted to a dataset under the assumption that the target variable matrix $\Y:=\left[\y_{1},\dots,\y_{n}\right]\in\bR^{m\times n}$ is incomplete. 
If the target variable $y_{k,i}\in\bR$ is larger than the detection limit~$u_{k}\in\bR$, then the numerical value of $y_{k,i}$ is available for fitting of MTTM; otherwise, the numeric value itself is unavailable. 
Denote by $\cE_{\text{h}}$ and $\cE_{\text{v}}$, respectively, the index sets of missing and visible entries, i.e. 
\begin{tsaligned*}
   & \cE_{\text{h}} := \left\{ (k,i)\in [m]\times [n] \,\middle|\, y_{k,i} \le u_{k} \right\}, 
   \\
   & \cE_{\text{v}} := \left\{ (k,i)\in [m]\times [n] \,\middle|\, y_{k,i} > u_{k} \right\}.  
\end{tsaligned*}

Let us see how the above-defined variables are defined for the toy data table given in Figure~\ref{fig:grphmdl}(c). This table has $n=6$ records. The first two columns contain the values of the target variables, leading to $m=2$ and $d=1$. For the first target values for the first three examples, $y_{1,1}$, $y_{1,2}$, and $y_{1,3}$ are below detection limit~$u_{1}$. Similarly for the second target variable, the values of the second and the fifth examples, $y_{2,2}$, and $y_{2,5}$ have been undetected. In this case, the index set of the missing entries in the matrix $\Y\in\bR^{2\times 6}$ are given by $\cE_{\text{h}}=\{ (1,1), (1,2), (1,3), (2,2), (2,5) \}$, and the other set $\cE_{\text{v}}$ contains the rest of the seven indices. 

To define the objective function for determination of the partial regression coefficient vector, a set of
\tsrev{%
probability
}%
density functions is introduced as follows: 
\begin{tsaligned}\label{eq:qv-m}
   \forall(k,i)\in\cE_{\text{v}}, \qquad
   q_{k,i}(t) = \delta(t-y_{k,i})\qquad
\end{tsaligned}
and 
\begin{tsaligned}\label{eq:qh-m}
\forall(k,i)\in\cE_{\text{h}}, \qquad
&
q_{k,i}(t) = 0 \quad\text{if }t>u_{k}, \\
&
q_{k,i}(t) > 0 \quad\text{if }t\le u_{k}. 
\end{tsaligned}
Let $\cQ_{k,i}:=\{\delta(\cdot-y_{k,i})\}$ for $(k,i)\in\cE_{\text{v}}$. 
For the other entries $(k,i)\in\cE_{\text{h}}$, 
define $\cQ_{k,i}$ as the set of 
\tsrev{%
probability
}%
density functions satisfying \eqref{eq:qh-m}. 
Denote the product space by
\begin{tsaligned}
\cQ_{\text{m}}
:=
\prod_{(k,i)\in[m]\times[n]}
\cQ_{k,i}
\end{tsaligned}
which corresponds to the space $\cQ_{\text{s}}$ for STTM. 
Recall that the log-likelihood function of STTM is the maximum of $\cF_{\text{s}}(\vthet_{\text{s}},q)$ with respect to $q$, a tuple of $n$ auxiliary functions, over the space $\cQ_{\text{s}}$. 
Combining the functions $\cF_{\text{s}}$ in Theorem~\ref{thm:sttm-is-max-F} is the basic idea adopted in this study to handle multiple censored target variables. 
The resultant function $\cF_{\text{m}}$ used for fitting of MTTM is defined as 
\begin{tsaligned}\label{eq:Fm-def}
& \cF_{\text{m}}(\vthet_{\text{m}},q)
:=
\sum_{(k,i)\in\cE_{\text{h}}}
   \cH[q_{k,i}] 
   +
   \sum_{i=1}^{n}
   \sum_{k=1}^{m}
   \int
   q_{1,i}(t_{1,i})
   \cdot
   \\
&
\dots\cdot
q_{m,i}(t_{m,i})
\log
\cN( t_{k,i} \,;\, 
\left<\va_{k},\vt_{\setminus k,i}\right>
+
\left<\vw_{k},\x_{i}\right>, \beta^{-1} ) 
d \vt_{i}
\end{tsaligned} 
where 
$\vt_{\setminus k,i}:=\left[t_{1,i},\dots,t_{k-1,i},t_{k+1,i},\dots,t_{m,i}\right]^\top$, 
$q:=(q_{k,i})_{k,i}\in\cQ_{\text{m}}$, and 
$\vthet_{\text{m}}:=(\va_{1},\dots,\va_{m},\vw_{1},\dots,\vw_{m},\beta)$. 
\tsrev{%
The regression coefficients can be regularized easily like ridge regression by adding 
\begin{tsaligned}\label{eq:reg-term}
   r_{\text{m}}(\vthet_{\text{m}})
   :=
   \frac{\lambda_{\text{reg}}}{2}
   \sum_{k=1}^{m}
   \left(
   \lVert\va_{k}\rVert^{2} 
   +
   \lVert\vw_{k}\rVert^{2} 
   \right)
\end{tsaligned}
to the objective function $\cF_{\text{m}}(\vthet_{\text{m}},q)$ with a small positive regularization constant $\lambda_{\text{reg}}$, which often reduces the generalization error and combats the multi-collinearity issue. 
In the experiments conducted in this study, $\lambda_{\text{reg}}:=10^{-3}$ is chosen. 
One may change the definition of $r_{\text{m}}(\vthet_{\text{m}})$ to change another regularization such as the lasso and the elastic net from the L2 regularization~\cite{Kashima-icpr2008}. 
}
To determine the value of the model parameters $\vthet_{\text{m}}$, MTTM solves the following optimization problem: 
\begin{oframed}
\textbf{MTTM fitting problem}
\begin{tsaligned}\label{eq:prob-max-Fm}
   \text{max }\quad& 
   \cL_{\text{m}}(\vthet_{\text{m}}) \qquad \text{wrt }\vthet_{\text{m}}, 
   \\
   \text{where }\quad& 
   \cL_{\text{m}}(\vthet_{\text{m}})
   :=
   \max_{q\in\cQ_{\text{m}}}\cF_{\text{m}}(\vthet_{\text{m}},q). 
\end{tsaligned}
\end{oframed}
Namely, the values of the partial regression coefficients in MTTM is determined by jointly optimizing the model parameters $\vthet_{\text{m}}$ and the auxiliary functions $q\in\cQ_{\text{m}}$. 
The details of the optimization algorithm is presented in the next section. 

\paragraph*{Imputation of values with censoring}
Once the optimal solution $(\vthet_{\text{m}}^{\star},q^{\star})$ is obtained, the values with censoring can be estimated with the statistical expectation $\bE_{q_{k,i}(t_{k,i})}[t_{k,i}]$. 
The following theorem indicates that the statistical expectation can be computed quickly. 
\begin{theorem-waku}\label{thm:mttm-q-trnorm}
   At the optimal solution of the maximization problem~\eqref{eq:prob-max-Fm}, say
   $(\vthet_{\text{m}}^{\star},q^{\star})$, 
   each entry in the set of the optimal functions $q^{\star}\in\cQ_{\text{m}}$, 
   denoted by $q_{k,i}^{\star}\in\cQ_{k,i}$, 
   is the 
\tsrev{%
probability
}%
   density function of the truncated normal distribution, i.e., 
   \begin{tsaligned*}
      \exists \mu_{k,i}^{\star}\in\bR, \, 
      \exists \sigma_{k,i}^{\star}\in\bR, \quad
      q_{k,i}^{\star} = f_{\text{TN}}(\cdot\,;\,\mu_{k,i}^{\star}, \sigma_{k,i}^{\star}, u_{k} ). 
   \end{tsaligned*}
   where 
   $f_{\text{TN}}(\cdot\,;\,\mu,\sigma,u)$ 
   is the density function of the truncated normal distribution: 
   \begin{tsaligned*}
      f_{\text{TN}}(x\,;\,\mu,\sigma,u)
   :=
   \begin{cases}
   \frac{\phi((x-\mu)/\sigma)}{\sigma \Phi((u-\mu)/\sigma)}
   &\,{ for }\, x\le u, 
   \\
   0 
   &\,{ for }\, x> u.
   \end{cases}
   \end{tsaligned*}
   Therein, $\phi$ and $\Phi$, respectively, denote the 
\tsrev{%
probability
}%
   density function and the cumulative distribution function of the standard normal distribution. 
\end{theorem-waku}
Proof is given in \tsrevtwo{Appendix~\ref{s:proof-thm-two-and-three}}. 
Letting $\widehat{u}_{k,i}^{\star}:=(u_{k}-\mu_{k,i}^{\star})/\sigma_{k,i}^{\star}$, 
the value of $y_{k,i}$ for $(k,i)\in\cE_{\text{h}}$ is estimated as 
\begin{tsaligned*}
   \bE_{q_{k,i}(t_{k,i})}[t_{k,i}]
   = \mu_{k,i}^{\star} - \sigma_{k,i}^{\star}\frac{\phi(\widehat{u}_{k,i}^{\star})}{\Phi(\widehat{u}_{k,i}^{\star})}
\end{tsaligned*}
which is the expectation of a random variable drawn from the truncated normal distribution~$f_{\text{TN}}(x\,;\,\mu_{k,i}^{\star},\sigma_{k,i}^{\star},u)$. 
\section{Fitting algorithm for MTTM}
\label{s:ca}
In the previous section, we described fitting MTTM to a dataset with censoring is performed by maximizing the functional $\cF_{m}(\vthet_{\text{m}},q)$. 
The authors employed the block coordinate ascent method to maximize the functional $\cF_{m}(\vthet_{\text{m}},q)$.
As shown in Algorithm~\ref{algo:ca-m}, the last step in each iteration (Line~\ref{line:thet-upd}) updates the model parameters~$\vthet_{\text{m}}$ with $q$ fixed. 
The other step (Line~\ref{line:qki-upd}) update an entry $q_{k,i}$ contained in $q$ for some $(k,i)\in\cE_{\text{h}}$ with all the other entries in $q$ and the model parameters frozen. 
\begin{algorithm}[t!]
\caption{
   Block coordinate ascent algorithm for maximizing $\cF_{\text{m}}(\vthet_{\text{m}},q)$.
   \label{algo:ca-m} }
\Begin{
      Initialize $\vthet_{\text{m}}$ and $q\in\cQ_{\times}$\;
      \For{$t:=1,2,\dots$}{
        \For{$(k,i)\in\cE_{h}$}{
          $q_{k,i} := \argmax_{q_{k,i}\in\cQ_{k,i}} \cF_{\text{m}}(\vthet_{\text{m}},q)$\; \label{line:qki-upd}
        }
        $\vthet_{\text{m}} := \argmax_{\vthet_{\text{m}}} \cF_{\text{m}}(\vthet_{\text{m}},q)$\;\label{line:thet-upd}
      }
   }
\end{algorithm}
As described in Algorithm~\ref{algo:ca-m}, each step poses a sub-problem for maximization of the functional~$\cF_{m}(\vthet_{\text{m}},q)$. The efficiency and the numerical stability of this block-coordinate ascent algorithm depends on the tractability of each sub-problem, which is ensured by the following theorem (Proof is given in \tsrevtwo{Appendix~\ref{s:proof-thm-two-and-three}}. ). 
\begin{theorem-waku}\label{thm:ca-m-upd}
  The solution of every step (Line~\ref{line:qki-upd} and Line~\ref{line:thet-upd}) in Algorithm~\ref{algo:ca-m} can be expressed in a closed form. 
\end{theorem-waku}
In what follows, the update formula of each step is presented.  

\subsection{Updating the density functions $q$}
\label{ss:upd-q}
Line~\ref{line:qki-upd} in Algorithm~\ref{algo:ca-m} is required to find the density function $q_{k,i}\in\cQ_{k,i}$ that maximizes $\cF_{m}(\vthet_{\text{m}},q)$ with the other functions in $q$ and the model parameters~$\vthet_{\text{m}}$ fixed. 
The function $q_{k,i}$ optimal to this sub-problem is expressed in a form of the truncated normal distribution: 
\begin{tsaligned}
  q_{k,i}^{\text{new}}(t_{k,i})
  =
  f_{\text{TN}}(t_{k,i};\mu_{k,i},\sigma_{k,i},u_{k}). 
\end{tsaligned}
Assuming the case of $k=1$, the optimal $\mu_{k,i}$ and $\sigma_{k,i}$ are described as follows. 
\begin{tsaligned*}
  \mu_{k,i}
  :=
  \frac{%
    \left<\vb_{1},\vW^\top\x-\vB_{\text{r}}^\top\overline{\y_{\text{r}}}\right>
  }{\lVert\vb_{1}\rVert^{2}},    
  \,
  \sigma_{k,i}
  :=
  \frac{1}{\sqrt{\beta}\lVert\vb_{1}\rVert}. 
\end{tsaligned*}
where $\vb_{1}\in\bR^{m}$ and $\vB_{\text{r}}\in\bR^{(m-1)\times m}$ are 
defined such that $\left[\vb_{1},\vB_{\text{r}}^\top\right] = \vI-\vA^\top$; 
the vector $\overline{\y}_{\text{r}}$ and the matrix $\vA$ are given as follows. 
The former is given by 
$\overline{\y}_{\text{r}}:=\left[ \overline{y}_{2,i},\dots, \overline{y}_{m,i} \right]^\top$ 
where 
$\overline{y}_{k',i}:=\bE_{q_{k',i}(t_{k',i})}[t_{k',i}]$. 
The $k'$th column in $\vA$ is defined as 
\begin{tsaligned*}
  \left[ a_{1,k'}, \dots, a_{k'-1,k'}, 0, a_{k'+1,k'}, \dots, a_{m,k'} \right]^\top. 
\end{tsaligned*}
The update rule for $q_{k,i}$ with $k\ne 1$ can be described by exchanging $k$ and $1$ in the description given here. 

\subsection{Updating the model parameters~$\vthet_{\text{m}}$}
\label{ss:upd-thet}
The sub-problem for Line~\ref{line:thet-upd} in Algorithm~\ref{algo:ca-m} is reduced to a ridge regression-like problem and thereby solved analytically. 
Define two vectors  
\begin{tsaligned*}
  \overline{\y}_{i}
  :=
  \left[ \overline{y}_{1,i},\dots, \overline{y}_{m,i} \right]^\top, 
  \quad
  \vv_{i}
   :=
   \left[ v_{1,i},\dots, v_{m,i} \right]^\top 
\end{tsaligned*}
each of which consists of the expectation with respect to the distribution $q_{k,i}$: 
\begin{tsaligned*}
\overline{y}_{k,i}:=\bE_{q_{k,i}(t_{k,i})}[t_{k,i}], 
\quad
{v}_{k,i}:=\bE_{q_{k,i}(t_{k,i})}[t_{k,i}^{2}]. 
\end{tsaligned*}
Denote by $\overline{\y}_{\setminus k,i}$ and $\vv_{\setminus k,i}$, respectively, the vectors obtained by excluding $k$th entry in $\overline{\y}_{i}$ and $\vv_{i}$. 
Let $\forall k\in[m]$, $\forall i\in[n]$, 
\tsrev{%
\begin{tsaligned}
   &
  \widetilde{\vw}_{k} := 
  \begin{bmatrix}
    \va_{k} \\ \vw_{k}
  \end{bmatrix}, 
  \qquad
  \widetilde{\x}_{k,i} := 
  \begin{bmatrix}
    \bar{\y}_{\setminus k,i} \\ \x_{i}
  \end{bmatrix}, 
  \\
  &
  \vG_{k}
  :=
  \text{diag}
  \left(
  \begin{bmatrix}
    \sum_{i=1}^{n}
    \vv_{\setminus k,i}
    \\ 
    \0_{d}
  \end{bmatrix}
  \right)
  +
  \lambda_{\text{reg}}\vI. 
\end{tsaligned}
One may set $\lambda_{\text{reg}}:=0$ if the L2 regularization term $r_{m}(\vthet_{\text{m}})$ is not added to the objective function. 
}%
With the density functions $q$ frozen, 
the value of the tuple $(\widetilde{\vw}_{1},\dots,\widetilde{\vw}_{m},\beta)$
maximizing the objective function $\cF_{\text{m}}(\vthet_{\text{m}},q)$ 
is expressed as
\begin{tsaligned*}
  &
  \widetilde{\vw}_{k}^{\text{new}}
  = 
  \left( 
    \vG_{k}
    + 
    \sum_{i=1}^{n}\widetilde{\x}_{k,i}\widetilde{\x}_{k,i}^\top 
  \right)^{-1}
  \sum_{i=1}^{n}\bar{y}_{k,i}\widetilde{\x}_{k,i}, 
  \\
  &
  \frac{1}{\beta^{\text{new}}}
  =
  \frac{1}{mn}
  \sum_{k=1}^{m}
  \left<\widetilde{\vw}_{k}^{\text{new}},\vG_{k}\widetilde{\vw}_{k}^{\text{new}}\right>
  \\
  &\qquad
  +
  \frac{1}{mn}  
  \sum_{k,i}
  (\bar{y}_{k,i}-
  \left<\widetilde{\vw}_{k}^{\text{new}},\widetilde{\x}_{k,i} \right>)^{2}
  +
  \frac{1}{mn}
  \sum_{k,i}
  v_{k,i}. 
\end{tsaligned*}
\tsrev{%
The new values of regression coefficients, $\vw_{k}^{\text{new}}$ and $\va_{k}^{\text{new}}$, are found in the vector $\widetilde{\vw}_{k}^{\text{new}}$ as 
\begin{tsaligned}
\widetilde{\vw}_{k}^{\text{new}} = 
\begin{bmatrix}
  \va_{k}^{\text{new}} \\ \vw_{k}^{\text{new}}
\end{bmatrix}.  
\end{tsaligned}
}
\newcommand{\underbarnashi}[1]{#1}
\begin{table}[t]
    \centering
\caption{
    \tsrev{%
    \textbf{Errors for imputation of censored data}. 
Three tables (a), (b), and (c) present the average values and standard deviations for RMSE at negative rates of 10\%, 20\%, and 30\%, respectively. In each cell, the left number is the average value, and the right number in a bracket is the standard deviation. The digits in bold indicate the smaller value within each row. The rightmost column indicates the P-value obtained by the paired $t$-test.  
    }%
  \label{tab:pval}}
  \tsrev{%
  \small
\begin{tabular}{l}
    (a) 
    \\
    \begin{tabular}{|c||c|c|c|} \hline
        Dataset & MTTM & STTM & P-value \\ \hline
        Indian & \textcolor{black}{\textbf{\underbarnashi{0.743}}} (0.178)  & \textcolor{black}{0.881} (0.126) & $2.048\times 10^{-5}$ \\ \hline
        NY Top & \textcolor{black}{\textbf{\underbarnashi{0.388}}} (0.068)  & \textcolor{black}{0.993} (0.080) & $2.316\times 10^{-63}$ \\ \hline
        NY Bottom & \textcolor{black}{\textbf{\underbarnashi{0.366}}} (0.081)  & \textcolor{black}{0.792} (0.091) & $4.211\times 10^{-44}$ \\ \hline
    \end{tabular}    
    \\
    \\
    (b)
    \\
    \begin{tabular}{|c||c|c|c|} \hline
        Dataset & MTTM & STTM & P-value \\ \hline
        Indian & \textcolor{black}{\textbf{\underbarnashi{0.978}}} (0.205)  & \textcolor{black}{1.180} (0.109) & $1.815\times10^{-8}$ \\ \hline
        NY Top & \textcolor{black}{\textbf{\underbarnashi{0.518}}} (0.070)  & \textcolor{black}{1.061} (0.065) & $9.600\times10^{-63}$ \\ \hline
        NY Bottom & \textcolor{black}{\textbf{\underbarnashi{0.477}}} (0.065)  & \textcolor{black}{0.957} (0.056) & $7.917\times10^{-62}$ \\ \hline
    \end{tabular}
    \\
    \\
    (c)
    \\
    \begin{tabular}{|c||c|c|c|} \hline
        Dataset & MTTM & STTM & P-value \\ \hline        
        Indian & \textcolor{black}{\textbf{\underbarnashi{1.057}}} (0.159)  & \textcolor{black}{1.324} (0.098) & $7.547\times10^{-17}$ \\ \hline
        NY Top & \textcolor{black}{\textbf{\underbarnashi{0.616}}} (0.065)  & \textcolor{black}{1.176} (0.061) & $4.733\times10^{-67}$ \\ \hline
        NY Bottom & \textcolor{black}{\textbf{\underbarnashi{0.595}}} (0.060)  & \textcolor{black}{1.053} (0.058)  & $2.849\times10^{-61}$ \\ \hline
    \end{tabular}
\end{tabular}
}%
\end{table}

\begin{figure*}[t!]
  \begin{center}
    \begin{tabular}{lll}
      (a) & (b) & (c)
      \\
      \includegraphics[width=0.3\textwidth]{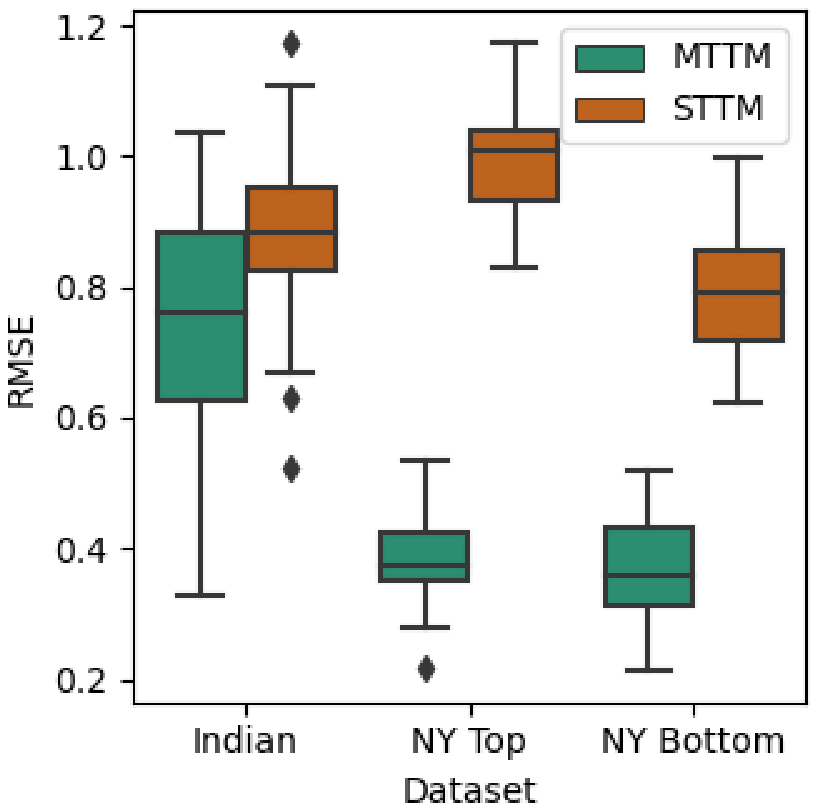}
      &
      \includegraphics[width=0.3\textwidth]{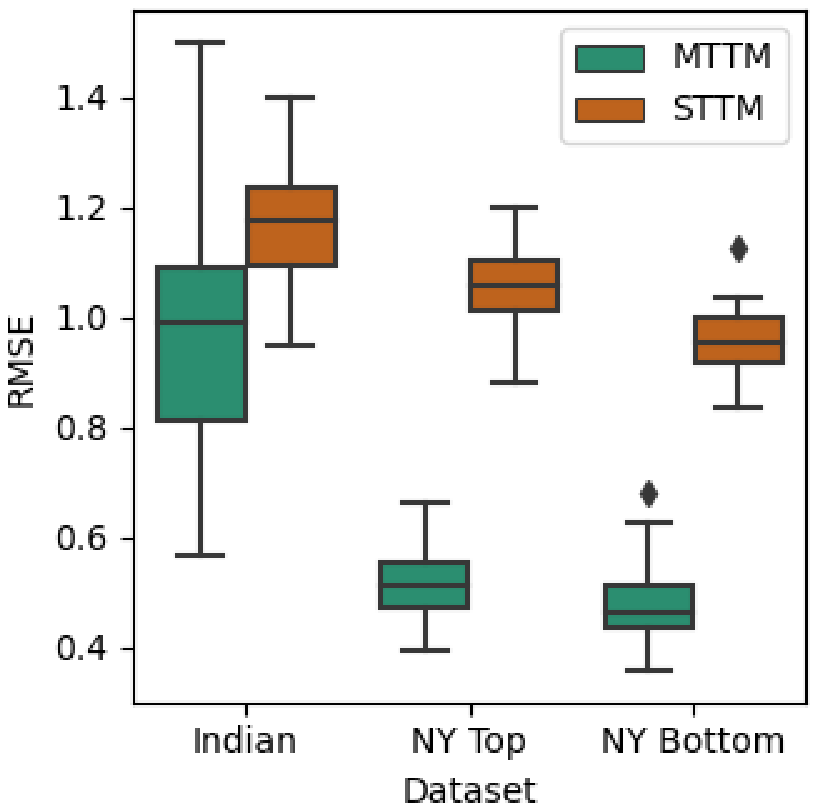}
      &
      \includegraphics[width=0.3\textwidth]{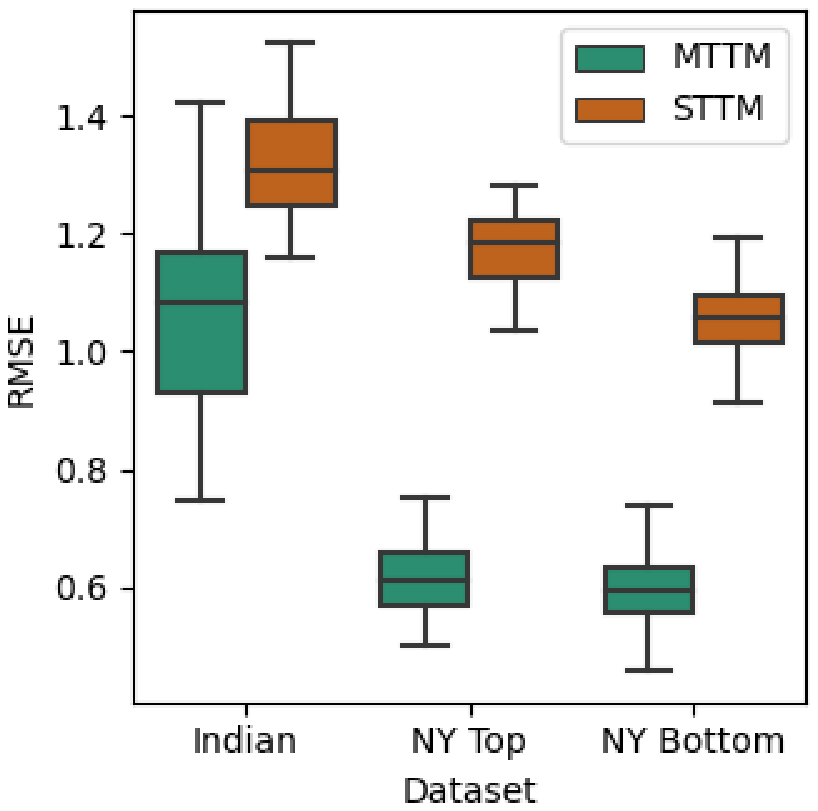}
    \end{tabular}
  \end{center}
  \caption{%
    \textbf{Estimation errors on three water quality datasets. }
    \tsrev{%
    Panels (a),(b),(c) show the box plot of RMSE for the case of negative rates 10\%, 20\%, and 30\%, respectively. Outliers are plotted as diamonds. 
    }%
    In every case, MTTM achieved much smaller estimation errors compared to STTM. 
    \label{fig:boxplot}
  }
\end{figure*}

\begin{table*}[t]
    \centering
\caption{
    \tsrev{%
    \textbf{Runtimes for 100 iterations}. 
    The unit of time is seconds. 
    The training set size $n$ was varied on a logarithmic scale from 10 to 1,000. The nine columns in the table show the computation times for nine different training set sizes. 
    }%
  \label{tab:runtime}}
  \tsrev{%
\begin{tabular}{|c||c|c|c|c|c|c|c|c|c|} \hline
    & 10  & 17 & 31 & 56 & 100 & 177 & 316 & 562 & 1,000 \\ \hline \hline
   Indian & 0.595 & 0.894 & 1.439 & 2.605 & 4.997 & 10.519 & 23.559 & 70.251 & 179.339\\ \hline
   NY Top & 0.457 & 0.912 & 1.926 & 3.438 & 5.619 & 14.087 & 31.023 & - & - \\ \hline
   NY Bottom & 0.635 & 0.867 & 1.728 & 3.125 & 5.705 & 11.399 & 26.520 & - & - \\ \hline
\end{tabular}
}%
\end{table*}
\begin{figure*}[t!]
  \begin{center}
    \begin{tabular}{ll}
      (a) & (b) 
      \\
      \includegraphics[width=0.4\textwidth]{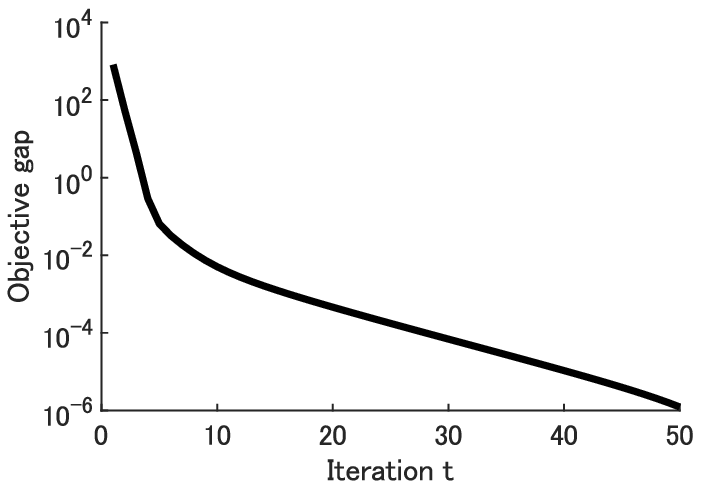}
      &
      \includegraphics[width=0.4\textwidth]{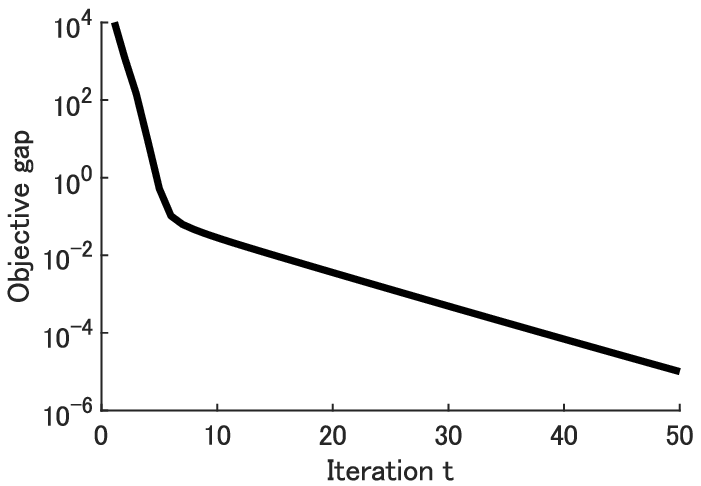}
    \end{tabular}
  \end{center}
  \caption{%
    \textbf{Convergence behaviors. }
    The algorithms with the cyclic order and the random order were examined. 
    The objective gaps monitored using the subset and the whole set of Indian were plotted in (a) and (b), respectively. 
    \label{fig:objgap}
  }
\end{figure*}

\section{Experimental results}
In this section, the performances of MTTM for estimating censored data is demonstrated (Subsection~\ref{ss:compl}). 
\tslong{%
Additionally, the convergence behaviors of the block-coordinate ascent method are reported (Subsection~\ref{ss:conv}).%
}%
\subsection{Estimation performances}
\label{ss:compl}
Computational experiments were conducted to compare the performance of MTTM with that of STTM. Three water quality datasets, Indian, NY Top and NY Bottom, were used in the experiment. Indian dataset contained 1,591 records collected from several sites in Indian rivers. The water quality variates are FC, TC, DO, BOD, pH, Cond, and Nitr. Abbreviations used here are given in \cite{KatKobOis19}.

The two datasets NY Top and NY Bottom were sampled from New York harbor and rivers near it. Measurements in NY Top and NY Bottom were from surface and bottom water, respectively. The two datasets contained 534 and 523 records, respectively. In both datasets, the water quality measurements were FC, TC, BOD, WMDO, pH, Cond, and WT. 


All the water quality measurements in these datasets are of indicators and are rarely censored. Artificial \emph{negative rates} were established for the first four variates in Indian, NY Top, and NY Bottom, 
\tsrev{%
where the negative rate means the percentage of data points below the detection limit, 
}%
in order to simulate these as actual pathogen concentrations. The sample size is set to $n=100$. For each randomly selected record, the detection limit was introduced so 
\tsrev{%
that 
}%
the negative rate coincides with the pre-defined value. Using this procedure, 50 samples were generated and used for assessment of imputation errors. 

\tsrev{%
Completion with STTM was done for each censoring column separately, following a typical approach in the water engineering field. 
Each censoring column is dealt as a target variate. The other columns were used as explanatory variates where the values under the detection limit were imputed in advance with the detection limit. This procedure was repeated with the target variate exchanged, to complete imputation of all missing values. 
}%

Figure~\ref{fig:boxplot} shows the box plots of the 50 root mean square errors (RMSE) obtained by MTTM and STTM. Panels (a), (b), and (c) plot the performances when the negative rates are set to 10\%, 20\%, and 30\%, respectively. 
\tsrev{%
The averages, the standard deviations and the P-values are shown in Table~\ref{tab:pval}. 
}%
For all cases, MTTM achieved significantly smaller RMSE compared to STTM. This suggests that estimating multiple columns simultaneously gains a great advantage over estimating them separately. 
\subsection{Time efficiency}
\label{ss:conv}
%
\tsrev{%
Here we demonstrate the time efficiency of the fitting algorithm of MTTM. Table~\ref{tab:runtime} reports the execution times of 100 iterations when applied to the three datasets with the training set size $n$ varied. 
The computer used to measure the execution times was equipped with 16GB memory and Intel(R) Core(TM) i7-10510U CPU @ 1.80GHz. 
As shown in Table~\ref{tab:runtime}, only three minutes accomplished 100 iterations even when the training set size was increased to 1,000. 
}

\tsrev{%
We next observed how accurate the solution at each iteration is to the optimum by monitoring the objective gap. The convergence speed was examined on the full set and a subset. The subset contained 100 records selected randomly from 1,591 records. Figure~\ref{fig:objgap} plots the objective gap between $\cF_{\text{m}}(\vthet_{m}^{\star},q^{\star})$ and $\cF_{\text{m}}(\vthet_{m}^{(t)},q^{(t)})$ against iterations $t$ where $(\vthet_{m}^{\star},q^{\star})$ was an optimal solution (estimated by running the algorithm for a very long time) and $ (\vthet_{m}^{(t)},q^{(t)})$ was the solution at the $t$th iteration. From the nature of the coordinate ascent algorithm, the objective gap was observed to be decreased monotonically. At the 17th iteration, the objective gap fell below $10^{-3}$ when the model was fitted to the subset. When the whole set was used, it took 27 iterations for the gap to fall below $10^{-3}$. 
Table~\ref{tab:runtime} and Figure~\ref{fig:objgap} suggest that the proposed method reaches near optimal values in practical time.
}
\section{Discussions}
\label{s:discuss}
%
%
\tsrev{%
For simplicity of discussion, the previous sections have been written with applications to left-censored water quality data with a constant detection limit for each variate, although the proposed algorithm has several straightforward extensions that can be useful for other domains with censoring. 
One of the extensions shall be presented here. Suppose that censoring occurs not only the left side but also the right side or both sides. 
Then, the available information of censored variate for $(k,i)\in\cE_{\text{h}}$ is described as 
\begin{tsaligned}
\begin{cases}
    y_{k,i} \le u_{k,i}
    &\quad
    \text{if $y_{k,i}$ is left-censored};  
    \\
    \ell_{k,i} \le y_{k,i}; 
    &\quad
    \text{if $y_{k,i}$ is right-censored};  
    \\
    \ell_{k,i} \le y_{k,i} \le u_{k,i} 
    &\quad
    \text{if $y_{k,i}$ is interval-censored}.  
\end{cases}
\end{tsaligned}
where $\ell_{k,i}$ and $u_{k,i}$ are a detection limit. 
With the change of available information, the definitions of $\cQ_{k,i}$ must be changed in a similar way. 
This change derives the following optimal density functions in Theorem~\ref{thm:mttm-q-trnorm}: 
\begin{tsaligned}\label{eq:q-opt-bothtrn}
    \exists \mu_{k,i}^{\star}\in\bR, \, 
    \exists \sigma_{k,i}^{\star}\in\bR, \quad
    q_{k,i}^{\star} = f_{\text{TN}}(\cdot\,;\,\mu_{k,i}^{\star}, \sigma_{k,i}^{\star}, \ell_{k,i}, u_{k,i} ). 
\end{tsaligned}
Therein, $f_{\text{TN}}$ is another truncated normal density function with four parameters redefined as 
\begin{tsaligned*}
f_{\text{TN}}(x\,;\,\mu,\sigma,\ell,u)
:=
\begin{cases}
 \frac{\phi((x-\mu)/\sigma)}
 {\left( \Phi((u-\mu)/\sigma) - \Phi((\ell-\mu)/\sigma) \right) \sigma }
 &\,\text{ if }\, \ell \le x\le u, 
 \\
 0 
 &\,\text{ otherwise}, 
\end{cases}
\end{tsaligned*}
where $\ell$ and $u$ are a constant that can take an infinite value. 
The update rule of $q$ is changed to the expression similar to \eqref{eq:q-opt-bothtrn}. 
The update rule of the model parameters $\vthet_{\text{m}}$ is unchanged. 
}

\tsrev{%
Thus, the proposed method possesses a potential applicable to a wide range of applications requiring analysis of censored data, since it can be easily extended to apply to right-censored and bilateral censored data as described in this section. 
}
\section{Conclusions}
\label{s:concl}
In this paper, a novel extension of the classical Tobit model was presented to handle multiple censored variables simultaneously by introducing multiple target variables. The extension is based on the fact that the second term in the likelihood function~\eqref{eq:L-s-def} for the classical Tobit model can be expressed with the minimum of a Kullback-Leibler 
\tsrev{%
divergence. 
}%
Based on this fact, multiple likelihood functions were combined to achieve the extended Tobit model. The main finding of this study is that if the block-coordinate ascent method is adopted for maximizing the reformulated objective function, each step can be expressed in a closed form, making the fitting algorithm numerically stable. Computational experiments using real-world microbiological water quality data empirically showed that the new Tobit model could estimate censored data with significantly higher accuracy than the classical Tobit model.

Data censoring occurs in many situations other than water quality measurement, as described in Section~\ref{s:related}. Evaluating the performance of the proposed model in other application areas is left to future work. 

\tsrev{%
Tobit model uses the normal noise model because of the least-squares estimation as its starting point, but there are also studies that fit other noise models to censored data~\cite{pmlr-v101-kohjima19a}. Exploring the practical effectiveness of these noise models combined with the proposed multi-target technique can be an interesting future challenge. 
}%

\section*{Acknowledgment}
This research was performed by the Environment Research and Technology Development Fund JPMEERF20205006 of the Environmental Restoration and Conservation Agency of Japan and supported by JSPS KAKENHI Grant Number 22K04372.
\appendix
%
\section{Derivation of \eqref{eq:min-KL-over-Q-is-neg-Z-in-main}: }
\label{s:proof-min-KL-over-Q-is-neg-Z-in-main}
\tsrevtwo{%
The right hand side of \eqref{eq:min-KL-over-Q-is-neg-Z-in-main}
is the Kullback–Leibler divergence from $p_{i}$ to $q_{i}$ 
minimized with respect to $q_{i}\in\cQ_{i}$, 
where $p_{i}:=\cN(\cdot\,;\,\left<\vw,\x_{i}\right>,\beta^{-1})$. 
To find the density function~$q_{i}\in\cQ_{i}$ minimizing the divergence, a Lagrangean function is defined as
\begin{tsaligned}
  \cL_{1}(q_{i},\eta)
  :=
  \text{KL}[q_{i}||p_{i}]
  +
  \left(
    1 - \int_{-\infty}^{u_{i}} q_{i}(x)dx 
  \right)
  \eta, 
\end{tsaligned}
where $\eta$ is the Lagrangean multiplier. 
Applying the variational method~\cite{bishop06}, the minimizer is 
expressed as $q_{i}(x) = p_{i}(x)/Z$ where
\begin{tsaligned}
  Z := 
  \int_{-\infty}^{u_{i}} p_{i}(t) dt. 
\end{tsaligned}
Substituting the minimizer into the Kullback–Leibler divergence, we have 
\begin{tsaligned}
  \text{KL}\left[\frac{p_{i}}{Z} \middle|\middle| p_{i}\right]
  &= 
  \int_{-\infty}^{u_{i}}
  \frac{p_{i}(t)}{Z}
  \log\frac{p_{i}(t)/Z}{p_{i}(t)}dt
  \\
  &=
  -\frac{\log Z}{Z}\int_{-\infty}^{u_{i}}p_{i}(t)dt = -\log Z
\end{tsaligned}
which is equal to the left right hand of \eqref{eq:min-KL-over-Q-is-neg-Z-in-main}. 
\qed
} 

\section{Proofs of Theorem~\ref{thm:mttm-q-trnorm} and Theorem~\ref{thm:ca-m-upd}}
\label{s:proof-thm-two-and-three}
\tsrevtwo{%
Both Theorem~\ref{thm:mttm-q-trnorm} and Theorem~\ref{thm:ca-m-upd} are based on the following fact: 
}
\begin{lemma-waku}\label{lem:upd-q}
  A probability density function $q_{k,i}^{\star}$ defined as 
  \begin{tsaligned}\label{eq:qki-star-in-lem:upd-q}
    q_{k,i}^{\star} \in \argmax_{q_{k,i}\in\cQ_{k,i}} \cF_{\text{m}}(\vthet_{\text{m}},q)
  \end{tsaligned}
  is that of the truncated normal distribution: 
  \begin{tsaligned}
    q_{k,i}^{\star}(t_{k,i})
    =
    f_{\text{TN}}(t_{k,i};\mu_{k,i},\sigma_{k,i},u_{k}). 
  \end{tsaligned}
  where $\mu_{k,i},\sigma_{k,i},u_{k}$ have been defined in Subsection~\ref{ss:upd-q}. 
\end{lemma-waku}
Note that $q_{k,i}$ in \eqref{eq:qki-star-in-lem:upd-q} is an entry of $q$. 

\tsrevtwo{%
\textbf{Proof of Lemma~\ref{lem:upd-q}: }
For simplicity of notation, let 
$\forall k_{1}\in[m]$, 
\begin{tsaligned}
  &
  \vt_{\setminus k_{1},i}
  :=
  \left[
    t_{1,i}, \dots, 
    t_{k_{1}-1,i}, t_{k_{1}+1,i}, \dots, t_{m,1}
  \right]^\top,
  \\
  &
  \text{ and }
  \quad
  q_{\setminus k_{1},i}(\vt_{\setminus k_{1},i})
  :=
  \prod_{k_{2}\ne k_{1}} q_{k_{2},i}(t_{k_{2},i}). 
\end{tsaligned}
The functional $\cF_{\text{m}}$ can be rearranged as 
\begin{tsaligned}\label{eq:0216a-A07}
  \cF_{\text{m}}(\vthet,q)
  &
  =
  \cH[q_{k,i}] 
  +
  \bE_{q_{k,i}(t_{k,i})}
  \left[
    \log f_{k,i}(t_{k,i})
  \right]
  +
  \text{const} 
\end{tsaligned}
where \text{const} denotes the terms having no dependency on $q_{k,i}$, and we have defined
\begin{tsaligned}
  & f_{k,i}(t_{k,i})
  :=
  \prod_{k'=1}^{m}
  \exp
  \bigg(
  \int 
  {q_{\setminus k,i}(\vt_{\setminus k,i})}
  \cdot
\\
&  
\log
\cN( t_{k',i} \,;\, 
\left<\va_{k'},\vt_{\setminus k',i}\right>
+
\left<\vw_{k'},\x_{i}\right>, \beta^{-1} ) 
d\vt_{\setminus k,i}
\bigg). 
\end{tsaligned}
By carefully rearranging the right hand side of the above equation, 
it can be observed that 
\begin{tsaligned}\label{eq:0216a-A09}
  \forall t_{k,i}\in(-\infty,u_{k}], \quad
  & f_{k,i}(t_{k,i})
  \propto
  \cN
  \left( t_{k,i}\,;\, \mu_{k,i}, \sigma_{k,i}^{2} \right). 
\end{tsaligned}
Define the Lagrangean function as
\begin{tsaligned}\label{eq:0216a-A10}
\cL_{2}(q_{k,i}) 
:= 
\cF_{\text{m}}(\vthet,q)
+
  \left(
    1 - \int_{-\infty}^{u_{k}} q_{k,i}(x)dx 
  \right)
  \eta, 
\end{tsaligned}
where $\eta$ is the Lagrangean multiplier. 
If the variational technique is applied to \eqref{eq:0216a-A10} combined with \eqref{eq:0216a-A07} and \eqref{eq:0216a-A09}, then we get 
\begin{tsaligned}
  q_{k,i}^{\star}(t_{k,i})
  =
  \frac{%
  \cN
  \left( t_{k,i}\,;\, \mu_{k,i}, \sigma_{k,i}^{2} \right)
  }{%
  \int_{-\infty}^{u_{k}}
  \cN
  \left( x\,;\, \mu_{k,i}, \sigma_{k,i}^{2} \right) 
  dx 
  } 
\end{tsaligned}
for $t_{k,i}\le u_{k}$. 
Thus, the optimal density function is obtained as 
$q_{k,i}^{\star} = f_{\text{TN}}(\cdot;\mu_{k,i},\sigma_{k,i},u_{k}).$ 
\qed 
}

\textbf{Proof of Theorem~\ref{thm:mttm-q-trnorm}: }
For $(k,i)\in\cE_{\text{h}}$, the density function $q_{k,i}^{\star}$ 
is of the truncated normal distribution. 
Combining this fact with Lemma~\ref{lem:upd-q} establishes Theorem~\ref{thm:mttm-q-trnorm}. 
\qed

\textbf{Proof of Theorem~\ref{thm:ca-m-upd}: }
Lemma~\ref{lem:upd-q} immediately derives the closed-form solution for Line~\ref{line:qki-upd} of Algorithm~\ref{algo:ca-m} as given in Subsection~\ref{ss:upd-q}. The solution of the sub-problem posed in Line~\ref{line:thet-upd} can be derived simply by setting $\frac{\partial\cF_{\text{m}}(\vthet,q)}{\partial\vthet}$
to zero. 
\qed

\renewcommand{\newblock}{}
\bibliographystyle{plain}

\end{document}